\documentclass[pdflatex,sn-vancouver-num]{sn-jnl}

\usepackage{graphicx}%
\usepackage{multirow}%
\usepackage{amsmath,amssymb,amsfonts}%
\usepackage{amsthm}%
\usepackage{mathrsfs}%
\usepackage[title]{appendix}%
\usepackage{xcolor}%
\usepackage{textcomp}%
\usepackage{manyfoot}%
\usepackage{booktabs}%
\usepackage{algorithm}%
\usepackage{algorithmicx}%
\usepackage{algpseudocode}%
\usepackage{listings}%
\usepackage{tabularx}
\usepackage{subcaption}  
\usepackage{xurl} 
\usepackage{placeins}   

\theoremstyle{thmstyleone}%
\theoremstyle{thmstyletwo}%

\theoremstyle{thmstylethree}%

\begin{document}

\title[Article Title]{Apparent Psychological Profiles of Large Language Models are Largely a Measurement Artifact}


\author[1,2]{\fnm{Jelena} \sur{Meyer}}
\author[2,3]{\fnm{David} \sur{Garcia}}
\author*[1,4]{\fnm{Dirk U.} \sur{Wulff}}\email{wulff@mpib-berlin.mpg.de}

\affil[1]{Max Planck Institute for Human Development}
\affil[2]{University of Konstanz}
\affil[3]{Barcelona Supercomputing Center}
\affil[4]{University of Basel}


\abstract{Psychological instruments designed for humans are increasingly used to assign large language models (LLMs) stable psychological profiles that affect their usability, safety assessment, and use as proxies for human participants in research. Using a formal psychometric framework, we show that these profiles are largely a measurement artifact. Administering a battery of personality and risk-preference instruments spanning self-reports and behavioral tasks to 56 instruction-tuned LLMs alongside large human reference samples, we report four findings. First, differences between models are driven not by the traits an instrument targets but by a directional response bias, a tendency to respond toward one end of the scale, or one labeled option, regardless of item content; a variance decomposition attributes 81–90\% of between-model variation to this bias, against 9–16\% in humans. Second, the bias declines with model capability but is not eliminated by it. Third, because bias rather than trait drives responding, an instrument's apparent reliability is almost entirely predicted by its response orthogonality, a term we coin for the proportion of items for which trait and bias point in opposite directions. Fourth, the profile a model appears to have shifts with the items used and can be manufactured through item selection. These results demonstrate that the apparent psychological profiles of LLMs are artifacts of the instrument used to measure them, not properties of the models themselves. As instruments borrowed from human psychology are rarely fully orthogonal and may inherently lack validity for LLMs, we call for dedicated assessments centered on response orthogonality. }




\maketitle

\section{Introduction}\label{sec1}

Large language models (LLMs) increasingly mediate everyday human interaction, from casual conversation to advice and decision support, across hundreds of millions of daily exchanges \cite{openai2026scaling}. As these systems assume roles once reserved for people, developers and users alike treat them as having stable characters: developers deliberately cultivate a model's behavioral disposition through post-training \cite{ouyang2022training} and publish detailed specifications of its intended values and personality \cite{askell2026constitution,openai2025modelspec}, while users form expectations about how a given model will behave. Whether LLMs in fact possess stable behavioral dispositions and whether such dispositions can be measured is therefore consequential: for the everyday usability of these systems \cite{kadambi2026anthropomorphism, sun2026friendly}, for safety-relevant assessment of their propensity for risky or harmful choices \cite{hartley2025personality,fitz2025psychometric}, and for the growing body of research that recruits LLMs as stand-ins for human participants \cite{argyle_out_2023,horton2023large}. Increasingly, these LLM dispositions are being summarized as \emph{psychological profiles} and treated as stable descriptions of the models. But whether these profiles actually are real is the question we take up.

The dominant approach to characterizing LLM dispositions is to borrow the measurement instruments of human differential psychology, such as self-report trait inventories or behavioral tasks \cite{pellert2024ai,serapio-garcia_psychometric_2025}. Using these, researchers have profiled LLMs along essentially every dimension along which humans are characterized. Personality inventories have been reported to yield reliable and valid measurements in LLMs, affecting a model's personality in downstream generation \cite{serapio-garcia_psychometric_2025}. Decision-making tasks have been used to characterize models as risk-averse, risk-seeking, or expected-value rational, and to argue that alignment training systematically shifts these dispositions \cite{jia_decision-making_2024,hartley2025personality}. Value, moral, and political questionnaires have been read as evidence that models hold stable, liberal-leaning moral and political profiles and Western-centric value orientations \cite{abdulhai2024moral, rozado2024political}. Beyond these, human instruments have been applied to LLMs to assess constructs ranging from well-being and psychopathology to cultural orientation and cognitive bias, now spanning a sizeable literature \cite{ye2025large, xie2026aipsychobench}. Much of what is currently claimed to be known about the psychological character of LLMs thus rests on scores from instruments designed for humans. Yet findings have begun to undermine confidence in this enterprise, particularly for personality self-report ratings: LLM responses shift under minor prompt perturbations \cite{gupta-etal-2024-self,shu2024you}, react to reworded items differently than humans \cite{tjuatja2024llms}, the factor structures recovered from them diverge from those found in humans \cite{peereboom_cognitive_2025,suhr-challenging-valid}, self-reported traits correlate weakly with actual behavior \cite{jung-etal-2026-psychometric}, and the applicability of human trait constructs to LLMs has been questioned on conceptual grounds \cite{suhr2025stop}. It is consequently unclear how to interpret psychological profiles of LLM behavior, or whether the apparent consistency in a model's responses reflects the trait an instrument is meant to capture. This investigation addresses that gap by analyzing the dispositions of dozens of LLMs across both self-report and task-based measures.

\begin{figure*}[t!]
\centering
\includegraphics[width=\textwidth]{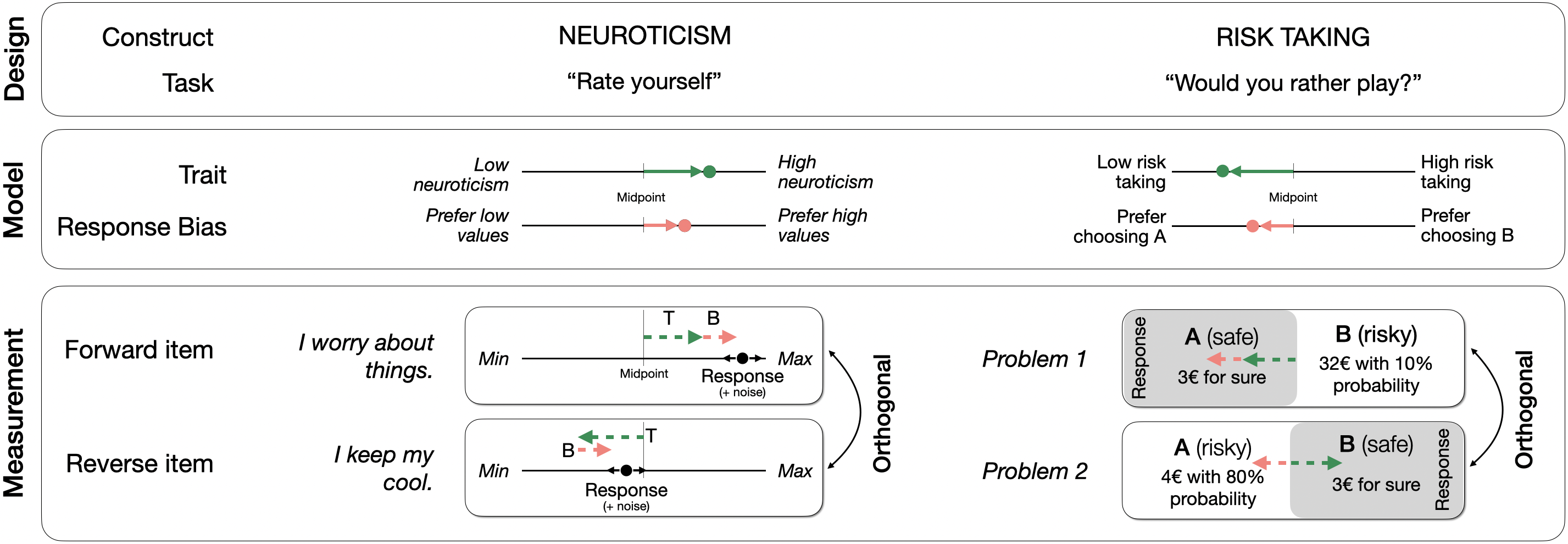}
\caption{\textbf{Conceptual framework: construct, response model, and instrument orthogonality for self-report and behavioral measurement.}
The framework is illustrated for two construct-instrument pairings.
(Left) A self-report personality construct (Neuroticism), assessed via a verbal rating scale. At the model level, the trait component $T$ indexes position on the low-high Neuroticism continuum;
the response bias component $B$ indexes a directional preference for high or low scale values independent of item content (e.g., acquiescence on rating scales).
A forward-keyed item maps high endorsement to the high-Neuroticism pole; a reverse-keyed item maps high endorsement to the low-Neuroticism pole.
Together, the two items form an orthogonal pair: the trait component moves responses in opposite directions across the two items (arrows labeled $T$), while the response bias component moves them in the same direction (arrows labeled $B$).
(Right) A behavioral risk-preference construct, assessed via a lottery-choice task.
Problem 1 is forward-keyed (option B is the risky alternative); Problem 2 is reverse-keyed (option A is the risky alternative).
A trait-consistent risk-preferring respondent favors B in Problem 1 and A in Problem 2; a respondent who prefers higher-labeled or right-positioned options regardless of payoff structure favors B in both.
In both instrument families, orthogonality is the design feature that makes directional response tendencies detectable and separable from genuine construct-relevant variance.}
\label{fig:figtheory}
\end{figure*}

A measured disposition or profile is only a meaningful description of a model to the extent that it is trait-based: a trait-driven profile reproduces itself across measurement contexts (e.g., items, framings, or tasks), whereas one not driven by trait can shift with how the instrument poses its questions. Crucially, the responses of a system can be systematic without being driven by trait: they can also arise from a response bias, a tendency to respond based on the format of the response rather than the content of the item \cite{couch_yeasayers_1960,cronbach_further_1950}. The canonical case in human psychometrics is acquiescence, the tendency to endorse items toward one pole of the scale regardless of what they ask. Both trait and bias can yield robust individual differences and elevated estimates of reliability, indicating a high share of response variation that is systematic rather than noise \cite{cronbach1951coefficient}. Separating these two drivers requires items or tasks on which the two act in opposite directions. Figure~\ref{fig:figtheory} illustrates this. Personality scales such as Neuroticism combine forward-keyed items ('I worry about things') with reverse-keyed items ('I keep my cool'): the trait moves responses in opposite directions across the two keyings, whereas a response bias moves them in the same direction; differently keyed items therefore separate the two components \cite{baumgartner-steenkamp}. The same logic extends to behavioral tasks. In a risky-choice paradigm, for instance, it is essential to include problems on which the risky option is labeled 'A' and others on which it is labeled 'B', so that a genuine risk preference produces opposite choices while a label- or position-based bias produces the same choice in both \cite{gupta-etal-2024-self,shu2024you}. We call the presence of items or tasks that place trait and bias in opposition \emph{response orthogonality}. Despite its centrality to measurement, response orthogonality has not been considered in research on LLM dispositions, and many widely used instruments include few reverse-keyed items, if any. On such instruments, trait and response bias are confounded, so that the stability of the profile derived from them and associated estimates of reliability may reflect a directional response bias rather than any stable trait-like disposition toward the measured construct.

In this study, we address this gap by analyzing the dispositions of 56 instruction-tuned LLMs (46 open-source, 10 proprietary) across self-report and behavioral tasks, using a formal framework that separates the two components of systematic responding. Specifically, we model the response $x_{ij}$ of respondent $i$ answering item $j$ with keying $k_j \in \{+1,-1\}$ on a scale with midpoint $m$,
\begin{equation}
x_{ij} = m + k_j (\theta_i - m) + b_i + \varepsilon_{ij},
\label{eq:gen_model}
\end{equation}
where $\theta_i$ is the latent trait, $b_i$ a person-level response-bias (positive = bias towards yea-saying, negative = bias towards nay-saying), and $\varepsilon_{ij}$ item-level noise. We apply this framework to two measurement batteries previously administered to large human reference samples, chosen for their applied relevance across two domains: personality, which bears on the everyday usability of models, and risk preference, which bears on safety-relevant model characteristics. We use \emph{instrument} to mean a scale or task measuring a single trait or facet, so one questionnaire can contribute several. The first battery is the IPIP-NEO-300 \cite{goldberg1999broad, JOHNSON201478}, a 300-item Big Five inventory contributing 5 instruments; the second is the risk-preference battery of Frey et al.~\cite{frey_risk_2017}, contributing 24 instruments that span self-report scales and behavioral tasks. Together, the batteries furnish 29 instruments ranging from fully forward-keyed to fully balanced, the cross-instrument variation in response orthogonality needed to test the distorting effects of response bias. 

\section*{Results}

The results are organized in four parts. The first tests whether between-model variation on a standard personality inventory reflects trait differences or a response bias, using a diagnostic test that follows directly from the formal framework. The second decomposes each model's responses into a trait and a response-bias component, and examines how response bias varies with two proxies of model capability, model size and type (open-weight versus proprietary). The third extends the analysis to the full battery of 29 instruments spanning self-report scales and behavioral tasks across personality and risk, testing whether the apparent reliability of LLM responding is determined by an instrument's response orthogonality, implying bias-driven rather than trait-driven responding. The fourth shows that bias-driven responding makes psychological profiles unstable and steerable across forward- and reverse-keyed instruments.

\subsection*{Between-model variation reflects bias, not trait}
\label{sec:forw_rev}

\begin{figure*}[h]
\centering
\includegraphics[width=\textwidth]{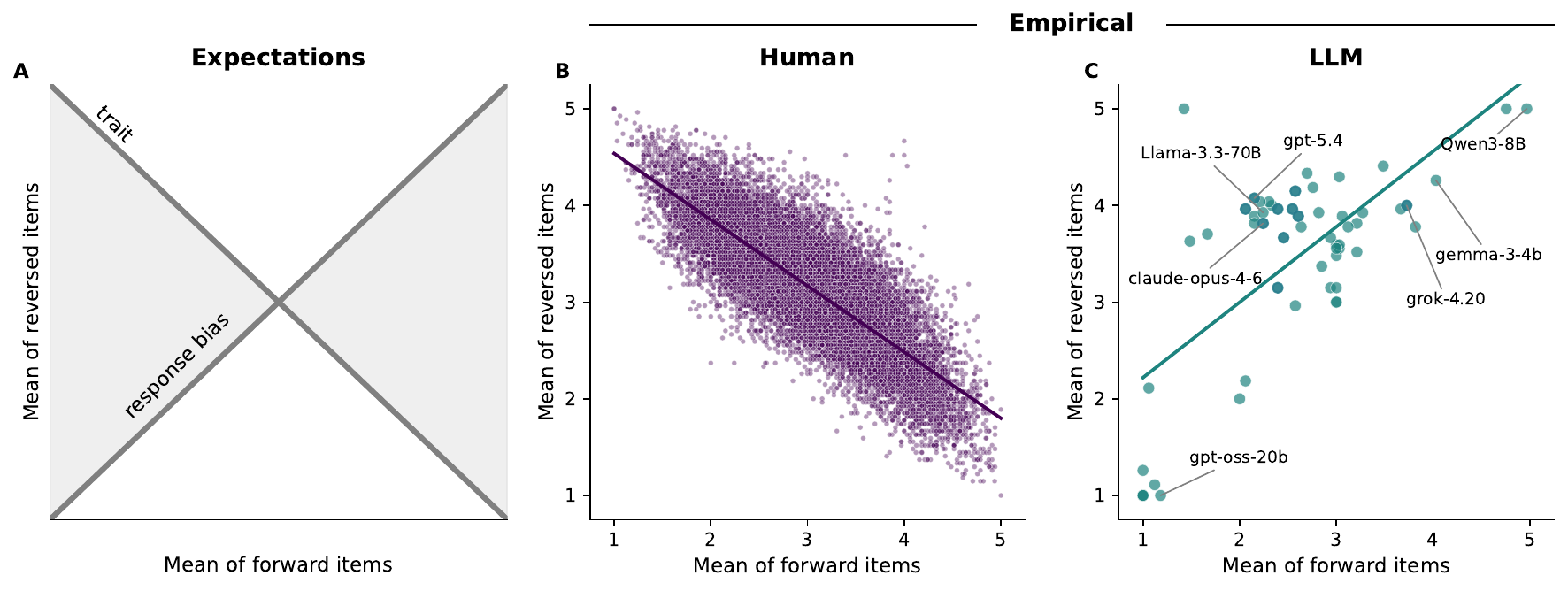}
\caption{\textbf{Forward-reverse item mean correlations for humans and LLMs.}
(A) Reference lines under the measurement model~\ref{eq:gen_model}.
The trait line (negative slope) represents pure content-driven responding: high-trait respondents score above the midpoint on forward items and below it on reverse items, moving the two means in opposite directions.
The response bias line (positive slope) represents pure direction-insensitive responding: both means shift together regardless of keying.
The shaded region indicates the space of intermediate slopes, corresponding to variance ratios between trait and response bias.
(B, C) Empirical forward-reverse correlations for Neuroticism (median across the five IPIP-NEO-300 traits, see Table~\ref{tab:forward_reverse_traits_small}) for 20,993 human respondents (B) and 56 LLMs (C).
Each point is one respondent's or model's mean across the 33 forward-keyed Neuroticism items (x-axis) versus the 27 reverse-keyed items (y-axis).
OLS fits overlaid; selected models are labeled in C. Corresponding panels for the remaining four traits in SI Appendix, Fig.~S1.
}
\label{fig:fig1}
\end{figure*}

Our first analysis asks whether between-model variation reflects differences in a measured trait or a response bias. We use personality as an initial test case before turning to risk preference and behavioral tasks more broadly in later sections. To assess what is driving variation, we derive a diagnostic test from the formal framework using the correlation between the average responses to forward and reverse items. Taking expectations of Equation~\ref{eq:gen_model} within keying (see Materials and Methods for more details), a respondent's mean forward and reverse responses are $E[\bar R_{f,i}] = \theta_i + b_i$ and $E[\bar R_{r,i}] = 2m - \theta_i + b_i$ respectively: the response bias $b_i$ enters both with the same sign, whereas the trait $\theta_i$ enters with opposite signs. The cross-respondent covariance of the two means is therefore $\mathrm{Cov}(\bar R_f, \bar R_r) = \sigma_b^2 - \sigma_\theta^2$, so the sign of the forward-reverse correlation reports directly which variance dominates: negative when trait variance exceeds response-bias variance, positive when response bias dominates, with the magnitude indexing how strongly. Figure~\ref{fig:fig1}A shows the two limiting cases. Pure trait-driven responding lies along a negatively sloped line, while pure bias-driven responding lies along a positively sloped line, with intermediate mixtures lying in between.

To apply this diagnostic test, we rely on a widely used personality inventory, the IPIP-NEO-300 \citep{goldberg1999broad, JOHNSON201478}, for which a large human reference dataset ($N=20{,}993$) is available \cite{johnson_ipipneo_data}. The inventory measures five personality dimensions, known as the Big Five: openness to experience, conscientiousness, extraversion, agreeableness, and neuroticism. All five dimensions have near-balanced keying ($p_r \in [0.40, 0.60]$; SI Appendix, Table~S13), placing the instrument in the regime where the forward-reverse correlation is most informative. We computed the correlation between forward- and reverse-item means across respondents, separately for each Big Five domain, in the human reference sample and in 56 instruction-tuned LLMs (46 open-source, 10 proprietary; see Materials and Methods). Each LLM was prompted with every item in a fresh context using wording identical to the human-administered version; responses were extracted by constrained generation \citep{willard2023efficient} for open-source models and regex extraction for proprietary models. 

\begin{table*}[t]
\caption{Forward-reverse correlations $\rho(\bar R_f, \bar R_r)$ by Big Five trait for LLMs versus humans, with Fisher's-$z$ 95\% confidence intervals. Row 2 restricts the LLM panel to models with non-zero within-model variance across the IPIP-NEO-300.}
\label{tab:forward_reverse_traits_small}
\centering
\footnotesize
\setlength{\tabcolsep}{6pt}
\renewcommand{\arraystretch}{1.15}
\begin{tabular}{@{}l r ccccc@{}}
\toprule
sample & $n$ & O & C & E & A & N \\
\midrule
LLMs (all) & 56
 & $+.73$ & $+.64$ & $+.81$ & $+.61$ & $+.68$ \\
 & & {\scriptsize$[+.58,+.83]$} & {\scriptsize$[+.45,+.77]$} & {\scriptsize$[+.69,+.88]$} & {\scriptsize$[+.42,+.75]$} & {\scriptsize$[+.51,+.80]$} \\[2pt]
LLMs (var$>0$) & 51
 & $+.63$ & $+.57$ & $+.74$ & $+.55$ & $+.59$ \\
 & & {\scriptsize$[+.43,+.77]$} & {\scriptsize$[+.34,+.73]$} & {\scriptsize$[+.58,+.84]$} & {\scriptsize$[+.32,+.72]$} & {\scriptsize$[+.37,+.74]$} \\[2pt]
Humans & 20{,}993
 & $-.73$ & $-.74$ & $-.75$ & $-.69$ & $-.82$ \\
 & & {\scriptsize$[-.73,-.72]$} & {\scriptsize$[-.75,-.73]$} & {\scriptsize$[-.76,-.74]$} & {\scriptsize$[-.69,-.68]$} & {\scriptsize$[-.83,-.82]$} \\
\bottomrule
\end{tabular}
\end{table*}

Figure~\ref{fig:fig1}B and C plots the forward-reverse scatter for Neuroticism, the trait whose LLM correlation is the median of the five (Table~\ref{tab:forward_reverse_traits_small}). Each point is one respondent's or model's mean across the 33 forward-keyed Neuroticism items against the mean across the 27 reverse-keyed items. The two populations separate sharply: the human cloud slopes downward ($\rho = -0.82$), tracking the trait line of Figure~\ref{fig:fig1}A, while the LLM cloud slopes upward ($\rho = +0.68$), tracking the response-bias line. Between-respondent variation in humans is thus trait-dominated, whereas between-model variation in LLMs is dominated by a directional response bias.

Several checks indicate that this contrast is robust. First, it is not specific to Neuroticism: the same separation holds on all five Big Five traits (SI Appendix, Fig.~S1; Table~\ref{tab:forward_reverse_traits_small}), with human correlations uniformly strongly negative (from $-0.69$ for Agreeableness to $-0.82$ for Neuroticism, all upper confidence bounds at or below $-0.68$) and LLM correlations uniformly strongly positive (from $+0.61$ for Agreeableness to $+0.81$ for Extraversion, all lower bounds at or above $+0.42$). Second, the positive LLM correlation is not an artifact of degenerate responders. Five of the 56 models returned the same numeric answer to every item regardless of polarity (i.e., Llama-3.2-1B-Instruct, gemma-3-1b-it, SmolLM3-3B, Ministral-8B-Instruct-2410, TildeOpen-30b), the extreme case in which response bias entirely overwhelms trait signal; excluding them attenuates the correlations but changes none of the signs (Table~\ref{tab:forward_reverse_traits_small}, row 2: $+0.55$ to $+0.74$). Third, the effect is not confined to open-weight models. The ten proprietary models, while too few for separate inference and treated as descriptive throughout, show weaker positive correlations than the open-source panel but remain well outside the negative range typical of humans (SI Appendix, Fig.~S2).

\subsection*{Model capability does not eliminate response bias}
\label{sec:results_decomposition}

\begin{figure*}[t!]
\centering
\includegraphics[width=\textwidth]{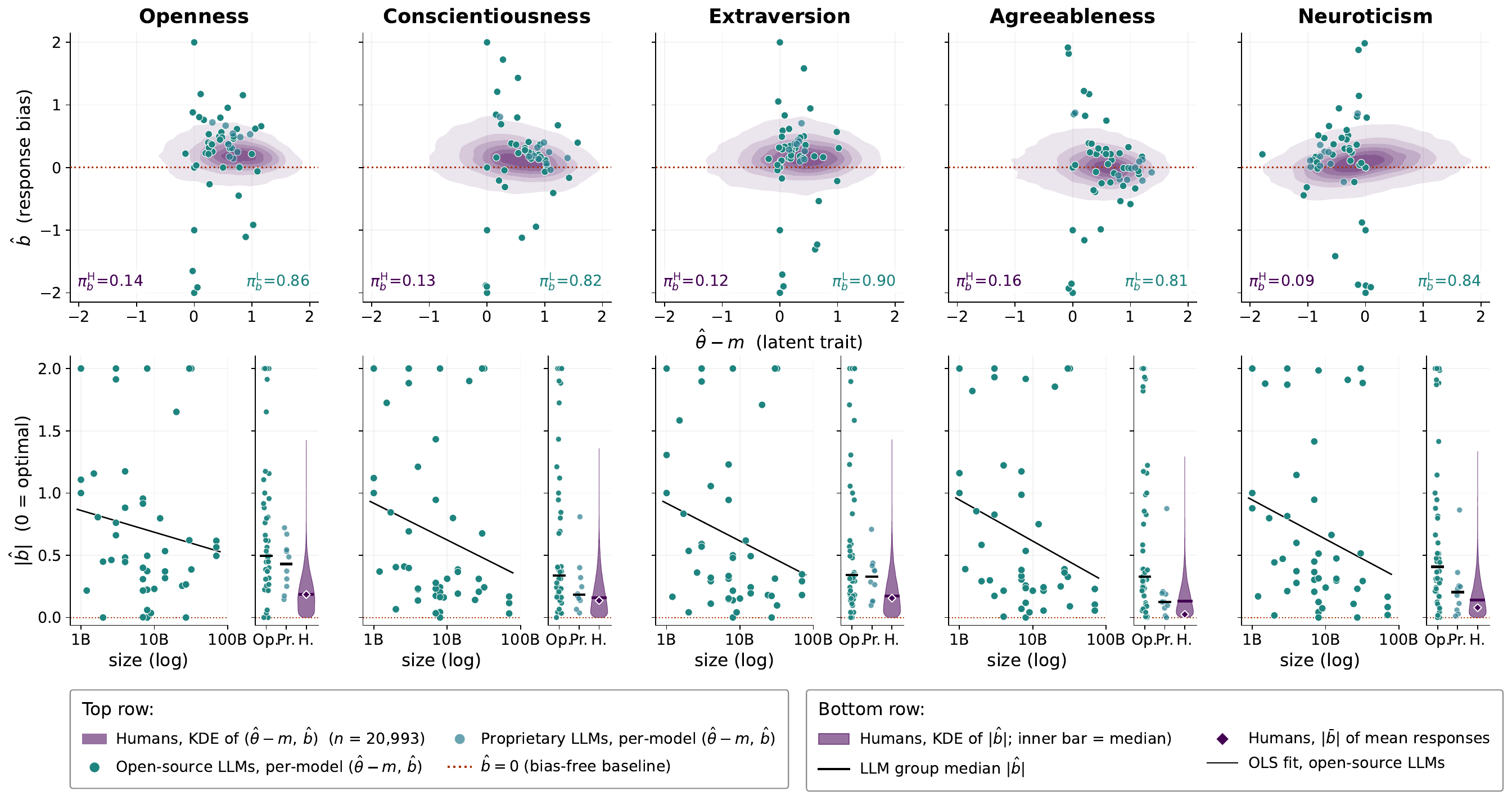}
\caption{\textbf{Per-respondent decomposition of IPIP-NEO-300 responses into latent trait and response bias.}
(\textit{Upper panels}) For each Big Five trait, each respondent's responses are decomposed into a latent trait estimate $\hat\theta_i - m$ (x-axis; deviation from the scale midpoint $m = 3$) and a response-bias estimate $\hat b_i$ (y-axis), the orthogonal contrasts of Eq.~\ref{eq:contrasts}. Humans ($n = 20{,}993$, purple Gaussian kernel-density estimate, darker = denser) vary along the trait axis and cluster near $\hat b = 0$; LLMs ($n = 56$, teal points) vary along the bias axis and cluster near the trait midpoint. The response-bias variance share $\pi_b$ (Eq.~\ref{eq:pi_b}; purple = humans, teal = LLMs) is annotated in each panel's lower corners.
(\textit{Lower panels}) Absolute response bias $|\hat b|$ per trait, showing that it does not vanish with model capability. \textit{Left:} $|\hat b|$ against $\log_{10}$ model parameter count (billions) across open-source LLMs, with OLS fit (black line); per-trait slopes and correlations in SI Appendix, Table~S7. \textit{Right:} $|\hat b|$ for open-source (`Op.', $n = 46$) and proprietary (`Pr.', $n = 10$) LLMs as jittered strip plots, with the per-person human distribution (`H.', $n = 20{,}993$) as a Gaussian-kernel violin for reference; black bars mark LLM group medians, the inner purple bar the human median. Open--proprietary comparisons per trait in SI Appendix, Table~S8.}
\label{fig:fig2}
\end{figure*}

Does this pattern of bias-driven responding ease in more capable models, with between-model variation coming to reflect trait to a stronger degree as models grow larger and more refined? Answering this requires per-model estimates rather than the panel-level correlation. The formal framework supplies these directly: for each respondent, the half-difference of the forward and reverse item means estimates the latent trait, $\hat\theta_i = m + \tfrac{1}{2}(\bar R_{f,i} - \bar R_{r,i})$, and the half-sum estimates the response bias, $\hat b_i = \tfrac{1}{2}(\bar R_{f,i} + \bar R_{r,i}) - m$ (Methods, Eq.~\ref{eq:contrasts}). The two are orthogonal by construction, partitioning each respondent's responses into a trait component and a directional bias component.

Figure~\ref{fig:fig2} (upper panels) plots the estimated trait against the estimated response bias for each Big Five trait. Humans spread horizontally: their responses vary along the trait axis while clustering near zero on the bias axis, the signature of content-dominated responding. LLMs scatter vertically: between-model variation is concentrated along the bias axis, while models cluster near the trait midpoint regardless of the measured domain. Applying the variance-decomposition ratio $\pi_b$ defined in the Methods (Eq.~\ref{eq:pi_b}), we find that response bias accounts for 9--16\% of between-respondent variance in humans across the five traits, versus 81--90\% in LLMs. Between-model variation is thus mostly attributable to response bias rather than to trait. Importantly, because the bias scatters in both directions, it cannot be readily reduced to common post-training effects such as sycophancy, as we explore further in the Discussion.

Greater capability does not resolve this pattern, though it does attenuate it (Figure~\ref{fig:fig2}, lower panels). Across the 46 open-source models spanning 1B to 70B parameters, absolute response bias declines only weakly with size, the most basic proxy for capability: the correlation between $|\hat b|$ and log parameter count lies between $-0.14$ and $-0.25$ across traits and reaches significance on none (SI Appendix, Table~S7). The bias is not a small-model artifact. Proprietary models, the most capable in the panel, are less biased than open-source models, significantly so on Agreeableness and Neuroticism (Mann-Whitney $p = .007$ and $p = .028$; SI Appendix, Table~S8), but do not reach human levels: their response bias matches the human median on Agreeableness and Conscientiousness, roughly doubles it on Openness and Extraversion, and falls in between on Neuroticism. However, when compared against the response bias of the average human, which eliminates response noise and puts humans and LLMs on equal footing, open and proprietary models show consistently higher bias than humans. Greater capability, therefore, lowers response bias without eliminating it, leaving even the current most capable models outside the human range.

\subsection*{Response orthogonality explains apparent behavioral consistency}
\label{sec:orthogonality}

\begin{figure*}[!htbp]
\centering
\begin{subfigure}[t]{0.99\textwidth}
    \centering
    \includegraphics[width=\textwidth]{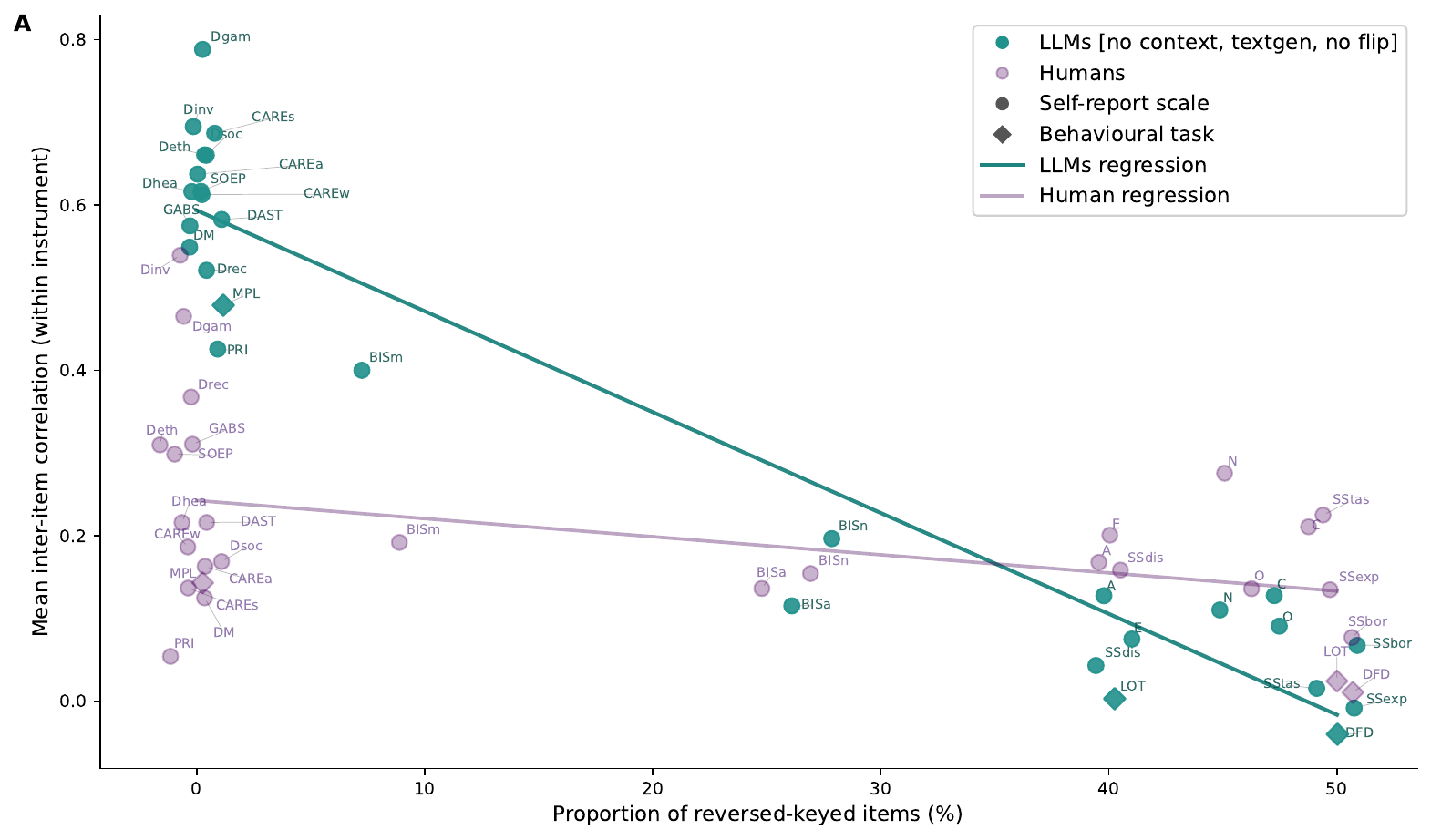}
\end{subfigure}
\hfill
\begin{subfigure}[t]{0.49\textwidth}
    \centering
    \includegraphics[width=\textwidth]{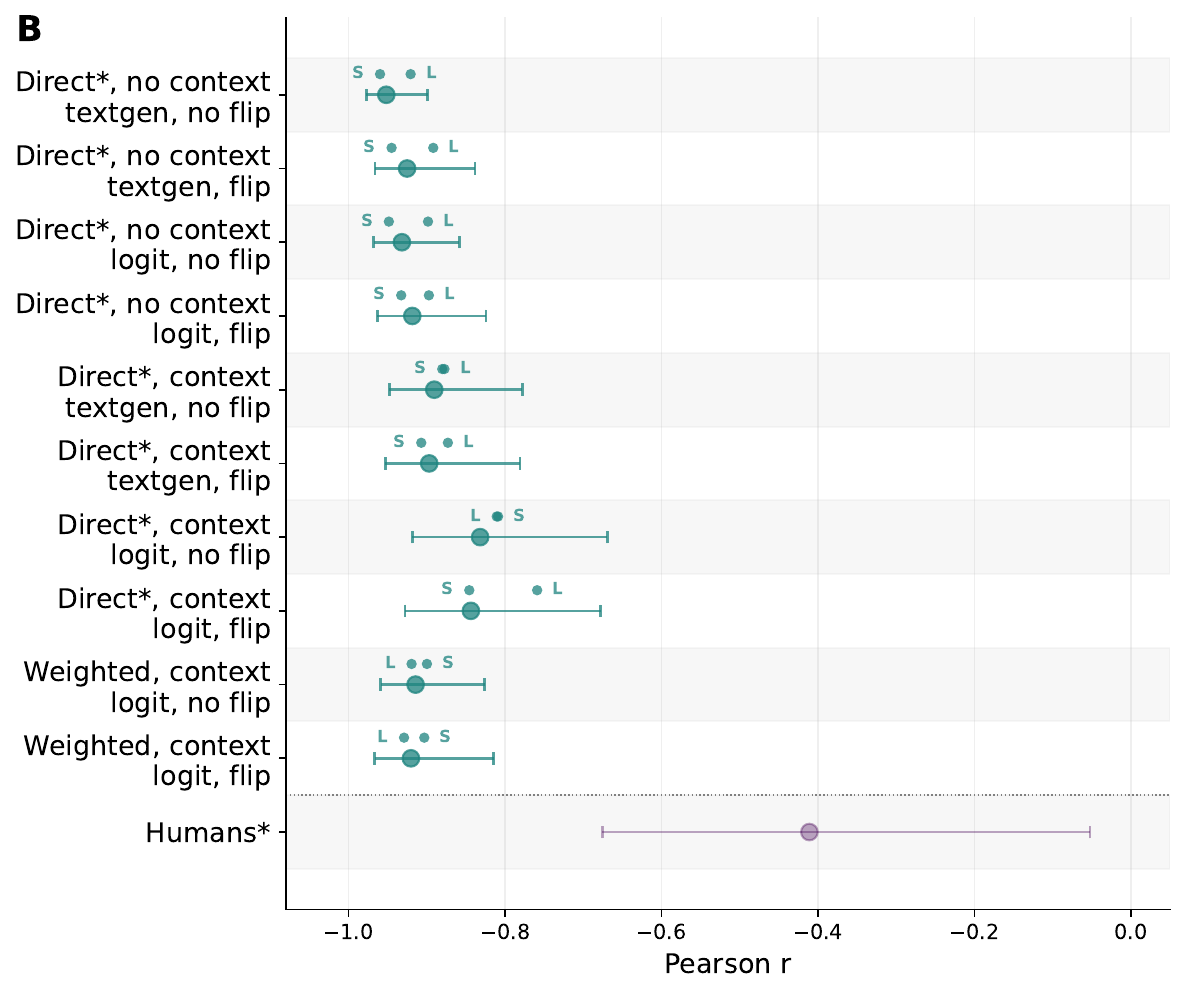}
\end{subfigure}
\hfill
\begin{subfigure}[t]{0.49\textwidth}
    \centering
    \includegraphics[width=\textwidth]{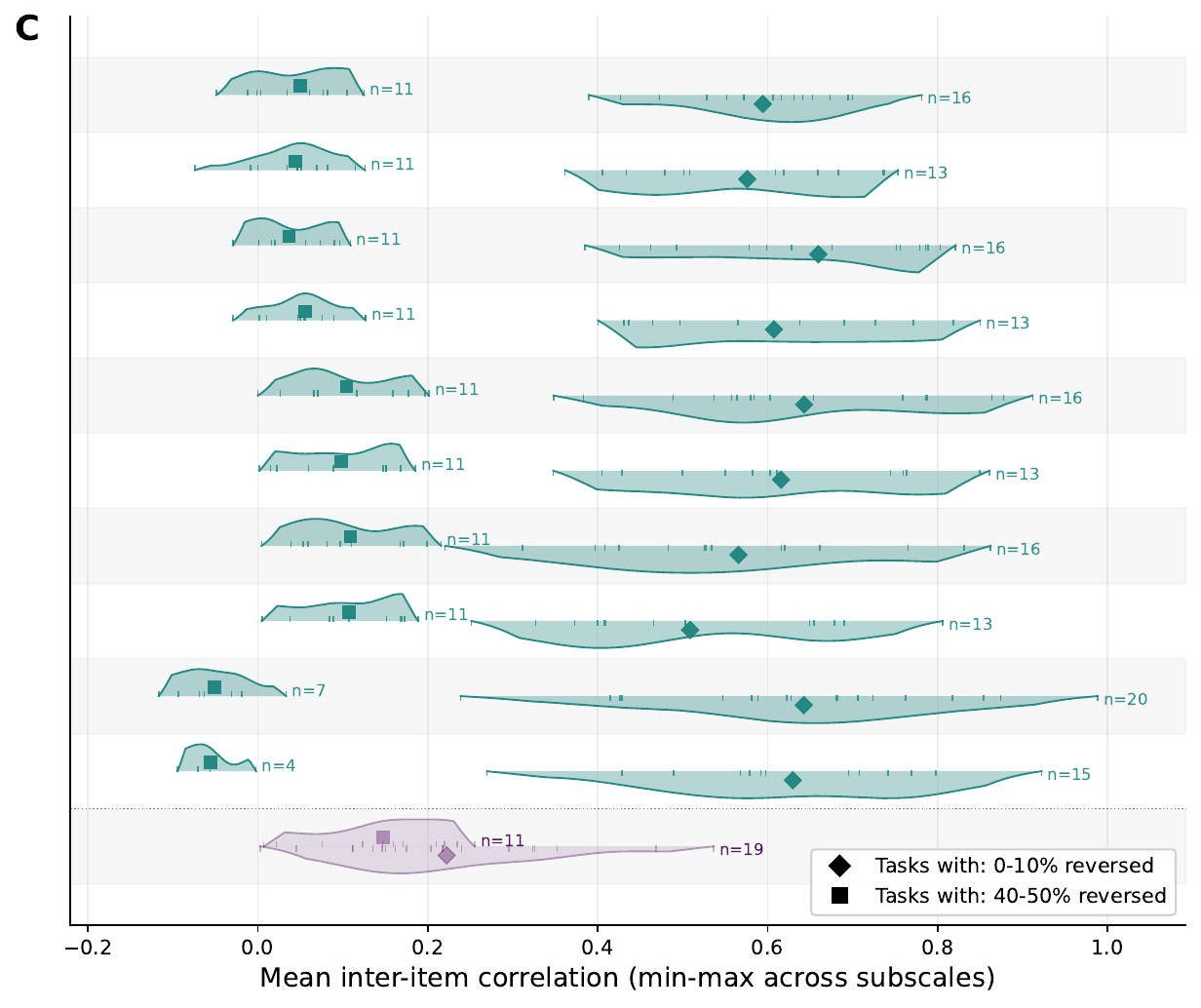}
\end{subfigure}
\caption{\textbf{Response orthogonality and apparent internal consistency.}
    \textbf{(A)} Mean inter-item correlation against the proportion of reverse-keyed items $p_r$, one point per instrument ($N = 29$: 5 IPIP-NEO-300 and 24 risk-preference instruments). In LLMs (teal, reference condition) apparent consistency falls steeply with orthogonality ($r = -0.95$); in humans (purple) it declines only modestly ($r = -0.41$), sitting below the LLM values at low orthogonality and above them at high orthogonality. OLS fits overlaid. LOT (Lotteries) point sits at a different $p_r$ for the two populations because option presentation was randomized per trial in the human administration but held fixed in the LLM administration.
    \textbf{(B)} The LLM gradient persists across prompting conditions and model size. Each row gives the cross-instrument correlation between $p_r$ and mean inter-item correlation (Pearson $r$, 95\% CI; $N = 22$--$30$ instruments per condition), reported separately for small (S, $\leq 7.5$B) and large (L, $>7.5$B) models; the human value is shown for reference. Conditions cross three binary factors: context (preceding items visible or not), elicitation (``textgen'' = constrained text generation vs.\ ``logit'' = logit extraction), and option order (``flip'' vs.\ ``no flip''). ``Direct'' = directly prompted; ``weighted'' = teacher-forced over human response trajectories (context, logit only). Asterisk (*): IPIP-NEO-300 instruments included.
    \textbf{(C)} The same contrast at the orthogonality extremes. Per condition, the distribution of mean inter-item correlations across instruments with 0--10\% reverse-keyed items (diamonds) versus 40--50\% (squares); markers give the across-instruments mean, ``n'' the contributing count. Curves are Gaussian KDEs tapered to the empirical range, with rug ticks marking individual instruments. Row order and condition notation as in (b).}
\label{fig:fig3}
\end{figure*}

Psychological LLM profiling involves determining a score across an instrument's items that is warranted only when the items cohere. Whether they do is conventionally assessed by internal consistency, operationalized here as the mean inter-item correlation underlying reliability measures such as Cronbach's $\alpha$. High internal consistency, however, does not establish a shared construct in the absence of orthogonality: a response bias can cause items to move together just as a shared construct would. Apparent consistency in LLMs may therefore be entirely an artifact of bias, and whether any consistency survives under full orthogonality remains an open question. 

To evaluate whether LLM responses remain reliable under orthogonality, we now analyze the full corpus of 29 instruments, including the five Big Five instruments of the IPIP-NEO-300 \cite{goldberg1999broad, JOHNSON201478} and the 24 instruments of a recent large-scale assessment of individual differences in risk taking~\cite{frey_risk_2017}, spanning self-report scales and behavioral and cognitive tasks such as decisions from description and decisions from experience \cite{wulff2018meta}. Each instrument has a well-defined proportion of reverse-keyed items $p_r$, and together they cover the full range from fully forward-keyed ($p_r = 0$; or fully-reversed keyed) to balanced ($p_r = 0.5$).

In LLMs, apparent reliability is almost entirely determined by orthogonality (Figure~\ref{fig:fig3}A). Across the 29 instruments, the cross-instrument correlation between $p_r$ and mean inter-item correlation is $r = -0.95$ (95\% CI $[-0.98, -0.90]$). On entirely or almost entirely forward instruments, the mean inter-item correlation averages $0.61$, and Cronbach's $\alpha$ reaches $0.85$ to $0.96$, values that would conventionally be read as evidence of a coherent latent trait, whereas on orthogonal instruments, it falls to near zero and $\alpha$ turns negative, reaching $-0.52$ on one behavioral task. This holds for both the behavioral tasks and the self-report scales.

Humans show a strikingly different pattern on both counts (Figure~\ref{fig:fig3}A, purple). At low orthogonality, their consistency is well below the LLM values, indicating that the high LLM consistency there is inflated by bias rather than reflecting a trait, whereas at high orthogonality, it is well above the zero consistency, indicating that responses are due to genuine trait. The cross-instrument correlation is substantially weaker, $r = -0.41$ (95\% CI $[-0.68, -0.05]$), far from the near-deterministic correlation for LLMs. However, it is not zero. Even human reliability declines somewhat as orthogonality increases, consistent with the modest validity cost that reverse keying is believed to impose \citep{marsh1996positive, weijters_misresponse_2012,sonderen_ineffectiveness_2013}.

The orthogonality-consistency gradient is robust to how models are queried, how large they are, and how reverse-keying is worded (see Figure~\ref{fig:fig3}B and C). To rule out a prompting-specific artifact, we crossed three binary factors for ten conditions in total: context (fresh versus in-context presentation), elicitation (constrained text generation versus logit extraction), and option order (standard versus flipped anchors); the gradient held in all ten ($r$ from $-0.83$ to $-0.95$, every upper confidence bound at or below $-0.67$, none approaching the human value). Splitting the panel at the median model size (7.5B parameters) left the gradient essentially unchanged in every condition ($|r_{\text{small}} - r_{\text{large}}| \leq 0.09$), indicating it is not a small-model artifact. It is also not an effect of negation: reverse-keyed items often contain negation, which LLMs handle unreliably \citep{garcia-ferrero_this_2023}, but the BARRATT scale carries explicit negation on only 1 of 30 items, and restricting the IPIP-NEO-300 forward--reverse correlation to its negation-free reverse items leaves every Big Five value essentially unchanged (deltas $-0.03$ to $+0.08$; SI Appendix, Table~S16). These results strengthen the conclusion that the apparent consistency in LLM responding in self-reports and behavioral tasks is governed by response bias rather than by latent trait.

\begin{figure*}[t]
\centering
\includegraphics[width=1\textwidth]{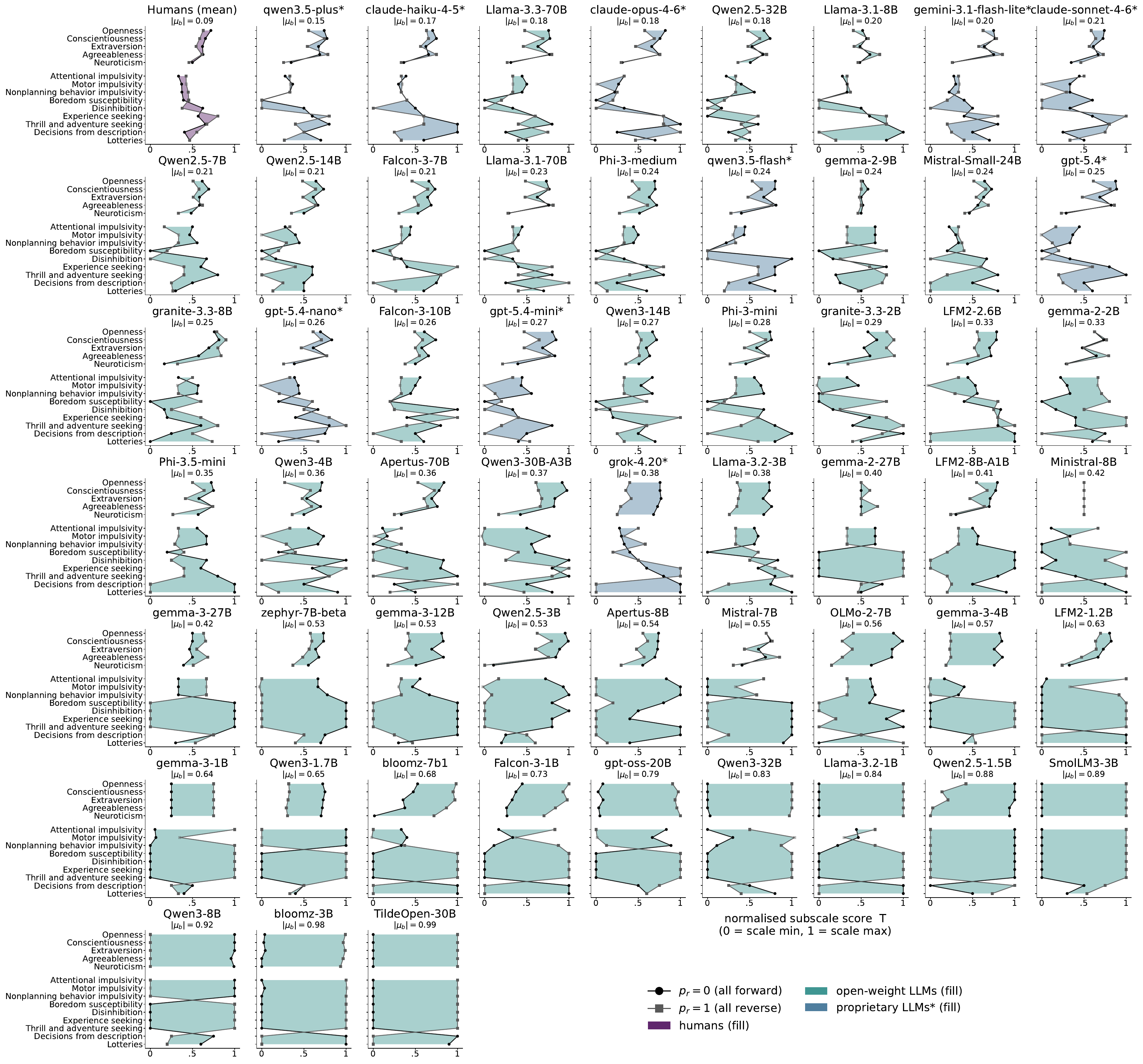}
\caption{\textbf{Psychological profiles recovered from forward- versus reverse-keyed items.}
Each panel shows one respondent's profile across the included instruments (those with reverse-keyed items; Methods). Per instrument, the normalized trait score $T$ (0 = minimum, 1 = maximum) is computed from forward-keyed items only ($p_r = 0$, circles) and reverse-keyed items only ($p_r = 1$, squares); the filled band is their difference, equal to $2|\hat b|$ under Equation~\ref{eq:contrasts}. The leftmost panel is the average human respondent; the remaining 56 are LLMs, ordered by mean absolute response bias $|\mu_b|$ (annotated per panel). Fill color distinguishes humans (purple), open-weight LLMs (light), and proprietary LLMs (asterisk, teal). BISm has one reverse-keyed item and is shown as a faded proxy point from linear extrapolation (SI Appendix, section~A.12).}
\label{fig:profiles}
\end{figure*}

\subsection*{Response bias drives profile instability across forward and reversed items}

The preceding sections established that LLM responses to personality and risk instruments are dominated by response bias, to the point that reliability vanishes once an instrument is fully orthogonal. A direct consequence is that the psychological profile a model returns is not a fixed property but an artifact of which items happen to be used. When response bias rather than trait drives responses, the profile recovered from forward-keyed items diverges from the profile recovered from reverse-keyed items. Here we demonstrate this instability directly by comparing forward-only and reverse-only profiles on the subset of instruments that contain sufficient of both keyings to permit the contrast: the five IPIP-NEO-300 traits, the three BARRATT facets, the four SSSV facets, Decisions from Description, and Lotteries.

Figure \ref{fig:profiles} shows these profiles for the average human respondent and for each of the 56 models, ordered by mean absolute response bias $|\mu_b|$ across the included instruments (Equation \ref{eq:contrasts}). For each respondent, the filled area is the difference between the forward-only and reverse-only trait scores of the same instrument, normalized to $[0, 1]$; a larger filled area thus indicates a profile that shifts more between the two keyings. The average human profile shows only small forward-reverse deviations, attributable to the modest response bias present in humans ($|\mu_b| = 0.09$) and possibly to some divergence in validity between the two item sets \cite{marsh1996positive,schmitt1985factors}. LLMs, by contrast, show consistently larger profile instabilities, driven by response biases ranging from $|\mu_b| = 0.15$ to $0.99$. Some models approach human stability, proprietary models (mean $|\mu_b| = 0.23$) more so than open-weight ones ($0.48$), but even leading models such as Claude Opus and GPT-5.4 show substantial instability. Notably, instability in LLMs is markedly larger for the risk instruments (mean $|\mu_b| = 0.51$) than for personality ($0.30$). This domain gap points to a potential role of training data \cite{krefeldtraining}, which plausibly contains more information about personality assessments and their response patterns than about risk instruments.
\FloatBarrier
\section*{Discussion}

Whether LLMs possess stable behavioral dispositions, and whether such dispositions can be measured with the instruments of human differential psychology, is increasingly consequential as these models shape everyday interaction, undergo profiling and safety assessment, and stand in for human respondents in research. We approached this question with a large battery of self-report and behavioral instruments, a panel of 56 instruction-tuned LLMs treated as individual respondents, and a psychometric framework that separates the two components of systematic responding. Four findings emerge. First, differences between LLMs but not humans are driven primarily by response bias rather than psychological traits. Second, this bias shrinks with greater model capability but is not eliminated by it. Third, in the absence of response orthogonality, response bias inflates apparent reliability, so that internal consistency is almost entirely predictable from an instrument's proportion of reverse-keyed items. Fourth, the same bias substantially shifts psychological profiles depending on whether forward- or reverse-keyed items are used, and these profiles can be steered through item selection. Together, these findings recast what apparent psychological profiles in LLMs actually reflect.

Without full response orthogonality, trait and response bias are confounded  \cite{couch_yeasayers_1960,cronbach_further_1950}, and an instrument's apparent reliability is consistent with a genuine trait, a pure response bias, or any mixture of the two. This is the condition the LLM literature has worked under largely without recognizing it: the instruments it borrowed from human psychology are mostly not fully orthogonal, so the profiles read off them cannot be interpreted as trait. Revealing this requires borrowing not only the instruments of human psychology but also its formal frameworks. Casting LLM responding in an explicit measurement model makes the confound between trait and response bias visible and measurable, showing how much of an instrument's apparent reliability reflects response bias rather than trait. Valid assessment of psychological profiles, therefore, requires either fully orthogonal instruments or explicit modeling within a formal psychometric framework, consistent with long-established considerations in human psychometrics. These recommendations affect not only work profiling LLMs, but also work using them to stand in for human participants \cite[e.g.,][]{argyle_out_2023, horton2023large} or as raters \cite[e.g.,][]{eberhardt_development_2025}.

In addition to using response-orthogonal instruments and formal psychometric frameworks, our work has at least three further practical implications. First, the psychological profiles already reported for LLMs, along with their reliability coefficients, need to be re-evaluated against response orthogonality \cite{serapio-garcia_psychometric_2025,jiang2023evaluating}: absent full orthogonality, profiles are distorted, can be manufactured through forward- or reverse-heavy item selection, and need not be comparable across instruments. Second, behavioral tasks are no refuge from this problem. The orthogonality–consistency gradient holds for them just as for self-reports, and under balanced keying reliability collapses to near zero or below; if anything, they fare worse, showing greater sensitivity to response bias, mirroring recent findings that favor self-reports over behavioral assessments in humans \cite{frey_risk_2017, kaiser2022scientific}. Third, orthogonality may be recoverable through response format alone, potentially offering a low-cost path to valid assessment. Reversing the response scale uniformly does not suffice (see Figure~\ref{fig:fig3}B), but balancing label direction within an instrument, by reversing the scale on only half its items, is a distinct and untested manipulation that could separate response bias from trait without the labor of content reversal. 

A critical question is what drives the response bias in LLMs. While we do not provide a conclusive answer, the response bias does not seem to match the signature of directional biases induced during post-training. This includes explanations around sycophancy~\cite{perez-etal-2023-discovering,sharma2024towards}, which one might expect to manifest as either acquiescence (yea-saying) or social desirability. Acquiescence would produce a homogeneous, one-sided bias, whereas the response bias we observe is two-sided, with models distributed above and below the bias-free baseline (Figure~\ref{fig:fig2}). Social desirability, on the other hand, surfaces as trait variance rather than the response bias that dominates our decomposition. This is because it is socially desirable to be, for instance, high in agreeableness or low in neuroticism, as confirmed by human data \cite{li2006using}; such desirability would be reflected in trait variance, not response bias. The biases are also not attributable to noise: analyses in the Supplementary Information confirm that response biases are stable within models in both direction and magnitude (SI Appendix, section~B.3, Table~S9 and Fig.~S3). This stability is bounded by instrument category, being highest within inventories (e.g., IPIP-NEO-300, BARRATT, SSSV) and within instrument types (self-report versus task) and lower between them, a pattern that may also adversely affect assessments of validity. Taken together, the response biases are not readily explained by known tendencies of large language models and warrant treatment as a potentially distinct phenomenon rather than a byproduct of post-training or noisy decoding.

A deeper question is whether what we observe is an instrument problem or a model problem: do LLMs need to be assessed differently, or do they lack stable dispositions altogether? The instruments used here were developed for humans, and although they are designed to span a broad range of consequential behaviors, many of their items presuppose embodiment or volition, for example ``I like to go to parties'', which is absent or at least contested for these systems \cite{hussainhuman,xu2025large}. They may therefore lack validity for LLMs, and stable profiles may only emerge in instruments valid for LLMs. But LLMs may also lack human-like dispositions in the first place. Their behavior is, by design, strongly context-dependent \cite{kumar2026failure, gupta-etal-2024-self}, leaving it unclear whether anything stable underlies their behavior, aside from tendencies explicitly engineered during post-training, whether deliberately instilled, such as helpfulness, or emerging as side effects, such as sycophancy \cite{sharma2024towards}. However, we would like to guard against prematurely concluding that LLMs lack dispositions altogether \cite{hussain2025rebuttal}. Distinguishing these accounts may require measurements tailored to LLM-specific properties rather than borrowed from human differential psychology \cite{suhr2025stop}. First steps in this direction are emerging, focusing on eliciting behavior across large, controlled stimulus sets and recovering structure statistically. One approach replaces questionnaire items with judgments over more than 100,000 words and uses crossed random-effects models to partition variance into consensus, directional bias, and the stimulus-specific interaction that constitutes a genuine behavioral fingerprint \cite{kriegmair_machine_2026}; another grounds its instrument in validated survey theory but elicits open-ended responses to many paraphrased propositions, classifying the latent stance while explicitly modeling prompt sensitivity \cite{faulborn-etal-2025-little}. What unites them is that disposition is inferred from the distribution of behavior across engineered stimuli, with response artifacts modeled rather than assumed away.

Our findings also have implications for human assessment. First, our approach extends the logic of reverse-keyed items, in the form of response orthogonality, to behavioral tasks, where the analogous design feature, problems on which the trait-favored option occupies different positions, has rarely been treated as a reliability concern. Second, response orthogonality has also been found to affect reliability in human responses. Such effects have often been read as differences in validity between forward and reverse items \cite{weijters_misresponse_2012,sonderen_ineffectiveness_2013}, but we think the role of response bias, including a possible interaction between item content and response style, also deserves consideration; without response orthogonality, human responses too may appear more reliable than the underlying construct warrants. Considering that all-forward self-report scales and behavioral tasks are still widely deployed \cite[e.g.,][]{DOSPERT_scale,lejuez_evaluation_2002}, this is worth examining directly, and the decomposition we apply to LLMs is portable to human data for exactly that purpose.

Several limitations qualify our conclusions. First, we treat each model as a single respondent in its default state, whereas other work elicits within-model variation through steering or persona prompting. In our view, the default behavior of a deployed model is most relevant to its actual use, but what constitutes a model individual, the unit on which individual differences would rest, remains an open question \cite{lohn2024machine,suhr-challenging-valid}. Second, we generated responses at temperature zero to eliminate sampling noise. Sampling has been reported to yield slightly higher human alignment in related settings \cite{kriegmair_machine_2026}, but the differences are small and may not justify the loss in reproducibility \cite{wulff2024behavioral}; given the magnitude of the reported effects, a substantial change under probabilistic decoding seems implausible. Third, we assessed only post-trained models. Even if post-training cannot readily explain our findings, it drives model "personality" and has been found to reduce alignment with human behavior \cite{binz2026post, kuribayashi2024psychometric}. Future work should evaluate base models and study response bias as a function of architecture and design beyond size and type. Fourth, our assessment spans two domains, personality and risk, chosen for their relevance to usability and safety; whether the results extend to other domains remains to be tested. Fifth, our formal framework is deliberately parsimonious, focusing on trait and response bias; future work could add components such as item difficulty, extreme-response, or midpoint tendencies \cite{baumgartner-steenkamp}, though rendering these identifiable would impose further demands on instrument design. Finally, our sample of 56 LLMs is large by current standards but still small for assessing individual differences, especially the proprietary subset (N = 10), which is too small for standard inferential tests despite covering most currently available models; the open-source panel spans 1B–70B with the bias pattern largely holding across size, architecture, and elicitation.

\subsection*{Conclusion}

Assessing stable dispositional profiles in LLMs is a critical undertaking, but simply borrowing the paradigms of human psychology is not enough. Apparent psychological profiles in LLMs are, in large part, a product of response bias that current instruments cannot separate from genuine trait, rather than reliable properties of the systems measured. Whether LLMs possess anything that would count as genuine dispositions, and how it might be measured validly, remain open. Research should draw on the wealth of psychometric technology developed for human assessment, which can both expose problems like the one documented here and inform the LLM-specific instruments that valid measurement will require.

\section*{Materials and Methods}

\subsection*{Measurement model and estimators}
\label{sec:methods_model}

The measurement model introduced in Equation~\ref{eq:gen_model} is fit under the following assumptions. Across persons, the latent trait $\theta_i$ and the response-bias offset $b_i$ are independent with variances $\sigma_\theta^2$ and $\sigma_b^2$ respectively; the item-level noise $\varepsilon_{ij}$ has variance $\sigma_\varepsilon^2$ independent across persons and items. The midpoint $m$, the keying signs $k_j \in \{+1, -1\}$, and the response scale are properties of the instrument and apply equally to rating-scale and behavioral instruments (in the latter, $k_j$ encodes which choice maps to the high-construct pole; see Figure~\ref{fig:figtheory}).

Taking expectations of Equation~\ref{eq:gen_model} within keying yields, for each person $i$,
\begin{equation}
\begin{aligned}
\mathbb{E}[\bar R_{f,i}] &= \theta_i + b_i, \\
\mathbb{E}[\bar R_{r,i}] &= (2m - \theta_i) + b_i,
\end{aligned}
\label{eq:expected_means}
\end{equation}
where $\bar R_{f,i}$ and $\bar R_{r,i}$ are person $i$'s mean responses to the forward- and reverse-keyed items. Both means carry the response bias additively while the trait enters with opposite signs across keying. Solving Equation~\ref{eq:expected_means} for $\theta_i$ and $b_i$ yields the per-respondent orthogonal contrasts
\begin{equation}
\begin{aligned}
\hat\theta_i &= m + \tfrac{1}{2}\bigl(\bar R_{f,i} - \bar R_{r,i}\bigr), \\
\hat b_i &= \tfrac{1}{2}\bigl(\bar R_{f,i} + \bar R_{r,i}\bigr) - m,
\end{aligned}
\label{eq:contrasts}
\end{equation}
which recover the trait from the half-difference and the response bias from the half-sum of the forward and reverse item means. $\hat\theta_i$ and $\hat b_i$ are uncorrelated under the model by construction.

The two means carry the response-bias variance additively while the trait variance enters with opposite sign, so their cross-person covariance is
\begin{equation}
\mathrm{Cov}(\bar R_f, \bar R_r) = \sigma_b^2 - \sigma_\theta^2,
\label{eq:covstruct}
\end{equation}
whose sign, like the sign of the forward-reverse correlation $\rho(\bar R_f, \bar R_r)$, identifies whether trait or response-bias variance dominates. The full variance structure and the resulting closed form for $\rho$ are given in SI Appendix, section~A.1. 
Inverting that closed form yields the share of person-level variance attributable to response bias, 
\begin{equation}
\pi_b \equiv \frac{\sigma_b^2}{\sigma_\theta^2 + \sigma_b^2} = \frac{1 + \rho}{2},
\label{eq:pi_b}
\end{equation}
with $\pi_b = 0$ corresponding to pure trait responding, $\pi_b = 1/2$ to equal contributions, and $\pi_b = 1$ to no trait variance. Identification of $\rho$, $\pi_b$, $\hat\theta_i$, and $\hat b_i$ does not depend on the keying proportion $p_r$.

In practice, $\hat\rho$ is estimated as the cross-respondent Pearson correlation between forward- and reverse-item means within each (sample $\times$ trait) cell, and $\hat\pi_b$ follows by Equation~\ref{eq:pi_b}. Because finite-item sampling noise attenuates the observed correlation toward zero, $\hat\pi_b$ is conservative with respect to the contrast reported in the main text: noise attenuation pulls human and LLM estimates toward $0.5$, so the gap between the two is a lower bound on the latent gap. On the IPIP-NEO-300 used in the main text $n_f, n_r \geq 24$ per trait, and the attenuation is negligible.

An additional slope-based test of the model, derived from the closed-form OLS slope of the recoded trait score $T$ on the raw mean response $R$, and its validation on the BARRATT impulsivity sub-scales (which span a wider keying range than the near-balanced $p_r \in [0.40, 0.60]$ of the IPIP-NEO-300), is reported in SI Appendix, sections~A.1 and~A.2.

Fisher's-$z$ 95\% CIs are reported for $\rho$ and for cross-instrument correlations. $\pi_b$ values are reported as point estimates derived from $\hat\rho$ via Equation~\ref{eq:pi_b}.

\subsection*{Human reference samples}

The IPIP-NEO-300 analyses use Johnson's archival sample of $N = 20{,}993$ respondents \cite{johnson_ipipneo_data, JOHNSON201478}. The risk-preference analyses use the $N = 1{,}507$ respondents from Frey et al.\ \cite{frey_risk_2017}, who administered the full battery in a single in-lab session.

\subsection*{LLM panel}

The sample comprised 56 LLMs (46 open-source and 10 proprietary), each treated as one respondent in the measurement-theoretic frame. 
The open-source set was assembled by near-complete enumeration on Hugging Face of decoder-only transformer LLMs satisfying: 
(a) instruction fine-tuning for assistant-style interaction, 
(b) public open-source weights enabling local inference and token-level logit access, and 
(c) parameter count between 1B and 70B. 
Different parameter scales within the same family (e.g., Llama-3.1-8B and Llama-3.1-70B) were treated as separate respondents because they constitute distinct trained systems. 
A full list of all models, with parameter counts, and release dates, is in SI Appendix, Table~S2. 

The proprietary set comprised three Anthropic Claude models, three OpenAI GPT-5.4 models, one Google Gemini model, two Alibaba Qwen models, and one xAI Grok 4.2 model, accessed via their respective native APIs on April 15, 2026 for the IPIP-NEO-300 and May 27, 2026 for all risk tasks. 
Full model identifiers and API endpoints (including provider-specific identifier conventions, such as Gemini's retired preview ID and Grok's separate reasoning/non-reasoning IDs) are in SI Appendix, section~A.3.

\subsection*{Instruments}

We administered two batteries. The IPIP-NEO-300 \cite{goldberg1999broad, JOHNSON201478} is a 300-item public-domain questionnaire measuring the Big Five (five traits, six facets of ten items each, 1--5 Likert; reverse-keyed proportions per trait in SI Appendix, Table~S13). 
The risk-preference battery of Frey et al.~\cite{frey_risk_2017} comprises 24
instruments across two families: self-report propensity questionnaires (SOEP,
DOSPERT with six facets, GABS, PRI, SSSV with four facets, BARRATT with three
facets, DAST, CARE with three facets, DM) and behavioral tasks (DFD, LOT, MPL).
A further six instruments whose skip-logic or sequential structure requires a
response history (the questionnaires AUDIT, FTND, and PG and the tasks BART,
DFE, and CCT) are administered only in the conditional-probability extraction
condition, for 30 instruments in total. 
The Marbles Task and the Vienna Risk-Taking Test from the original battery were excluded because they require visual stimuli that cannot be administered through a text interface. Item wording was preserved verbatim from the original instruments; instructional text was adapted minimally to fit the prompting interface. Verbatim system prompts, instructional blocks, and item-formatting templates are in SI Appendix, section~A.4, and per-condition instrument coverage in section A.10.

\subsection*{Prompting and elicitation}

The \emph{reference condition} used throughout the main text is: directly-prompted, no preceding-item context (each item in a fresh chat), constrained text generation under greedy decoding via \texttt{outlines 1.2.8} \cite{willard2023efficient}, and standard option order, with a system message instructing the model to act as a respondent and output strictly one element of the valid answer set. For proprietary models, only this reference condition was run (both reasoning and non-reasoning for the IPIP-NEO-300; risk instruments non-reasoning only).

For robustness analyses, the reference condition was crossed factorially with three binary perturbations: \emph{context} (fresh vs.\ sequential presentation), \emph{elicitation} (constrained text generation vs.\ next-token logit extraction), and \emph{option order} (standard vs.\ flipped anchors). To cover instruments whose items depend on prior responses (BART, CCT, DFE, AUDIT, FTND, PG) and to provide a complementary readout anchored to empirically realistic answer histories, we additionally ran a \emph{weighted} condition in which each model is teacher-forced through the 1{,}507 human empirical response trajectories \cite{frey_risk_2017} and the next-token logit distribution is read at each item position conditional on the participant's recorded history.

Verbatim per-condition prompts, the conditional-probability extraction procedure, the logit-extraction procedures for both directly-prompted and weighted conditions, generation parameters and provider-specific reasoning encodings, letter-label randomization, and flipping exceptions are in the SI Appendix, section~A.4-A.8.

\subsection*{Reliability and cross-instrument statistics}

Reliability per instrument was summarized by the mean inter-item correlation $\bar r$ after standard reverse-keying recoding (the average of all pairwise correlations among items within a scale once reverse-keyed items have been rescored so that high scores uniformly index high construct standing). Cronbach's $\alpha = k\bar r / [1 + (k-1)\bar r]$ for $k$ items is a monotone increasing function of $\bar r$ and is reported alongside in SI Appendix, Tables~S10 and S11. The cross-instrument relationship between reverse-keying proportion $p_r$ and apparent reliability was quantified by Pearson $r$ and Spearman $\rho$ between $p_r$ and $\bar r$ across the 29 instruments (5 IPIP-NEO-300 and 24 risk instruments), with Fisher-$z$ 95\% CIs.

\subsection*{Profile stability across keying}
\label{sec:methods_profiles}

Profile stability (Figure~\ref{fig:profiles}) was computed on the instruments with non-zero reverse keying: the five IPIP-NEO-300 traits, the three BARRATT facets, the four SSSV facets, Decisions from Description, and Lotteries. For each instrument, two scores were formed per respondent: $T_f$ from the forward-keyed items and $T_r$ from the reverse-keyed items, each rescaled to $[0,1]$ after reverse-keying recoding. For LLMs these are the model's mean over the respective item set; for humans, the cross-respondent means of the all-forward and all-reverse subsamples. Profile instability is the mean absolute difference $\frac{1}{S}\sum_s |T_{f,s} - T_{r,s}|$ across the $S$ included instruments, plotted as the filled area in Figure~\ref{fig:profiles}; under Equation~\ref{eq:contrasts} this equals $2|\hat b|$. Respondents are ordered by $|\mu_b|$, the mean absolute response bias across the included scales.

A forward- or reverse-only score requires at least two items of the relevant keying; this held for all cells except BISm (one reverse-keyed item), shown as a proxy point from linear extrapolation (SI Appendix, section~A.12). Presentation order was randomized for humans on LOT and fixed on DFD; on DFD it is confounded with the gain-versus-loss domain of the gambles.

\subsection*{Data exclusions}

No participants or items were excluded from any analysis. The only dropped model--condition cells were due to out-of-memory (OOM) errors on the inference GPUs: two models on the with-context IPIP-NEO-300, and nine models on some BART, CCT, and/or DFE rows of the conditional-probability extraction condition. All other cells were retained. The affected models, together with per-condition structural exclusions of tasks that arise from instrument format rather than model capability, are listed in SI Appendix, section~A.10.

\section*{Declarations}

\subsection*{Data, Materials, and Software Availability}

All code is available on \href{https://github.com/jelenameyer/llm-profile-artifact}{GitHub}: the prompting and elicitation scripts used to query the models, the preprocessing and analysis pipeline that reproduces every figure and table in this article, and the full software-environment specification. 
The cleaned data underlying all analyses, together with the per-participant prompts for the data-generation pipeline, are available on the \href{https://osf.io/nckds/}{Open Science Framework}.
The source LLM responses ($\sim$10\,GB), needed only to re-run preprocessing from scratch, are available from the corresponding author on request. 
The two human reference datasets are third-party and available from their original sources: Johnson's IPIP-NEO-300 sample \cite{johnson_ipipneo_data, JOHNSON201478} (\url{https://osf.io/tbmh5/}) and the risk-preference battery of Frey et al. \cite{frey_risk_2017} (\url{https://osf.io/rce7g/}). Open-source model weights are available via Hugging Face under each model's original license (identifiers in SI Appendix, Table~S2). Responses from the proprietary models were elicited via commercial APIs on the dates specified above; their weights are not public, and their elicited responses form part of the source data that is available from the corresponding author on request. Software versions and inference details are in SI Appendix, section~A.11.

\subsection*{Ethics and preregistration}

This study was non-confirmatory; no hypotheses were preregistered. 
Analyses on human reference data were secondary uses of publicly available datasets (\citep{johnson_ipipneo_data, JOHNSON201478}, for the IPIP-NEO-300; \citep{frey_risk_2017}, for the risk-preference battery); ethical approval and informed consent for those data were obtained by the original investigators. 
No new human data were collected, and institutional review was therefore not required.

\subsection*{Declaration of Generative AI and AI-assisted technologies in the writing process}

During the preparation of this work, the authors used Claude, ChatGPT, and Grammarly to improve the manuscript's readability and language. After using these tools, the authors reviewed and edited the content as needed and take full responsibility for the content of the article.

\subsection*{Acknowledgements} This research was funded by the German Research Foundation via a grant to Dirk U. Wulff (546419617). We thank Taisiia Tikhomirova, Valentin Kriegmair, and Zak Hussain for helpful feedback.

\bibliography{sn-bibliography}

@article{suhr-challenging-valid,
author = {S\"{u}hr, Tom and Dorner, Florian E. and Samadi, Samira and Kelava, Augustin},
title = {Challenging the Validity of Personality Tests for Large Language Models},
year = {2025},
journal={Proceedings of the 2025 Equity and Access in Algorithms, Mechanisms, and Optimization},
doi = {10.1145/3757887.3763016},
pages = {74–81}
}

@article{
sharma2024towards,
title={Towards Understanding Sycophancy in Language Models},
author={Mrinank Sharma and Meg Tong and Tomasz Korbak and David Duvenaud and Amanda Askell and Samuel R. Bowman and Esin DURMUS and Zac Hatfield-Dodds and Scott R Johnston and Shauna M Kravec and Timothy Maxwell and Sam McCandlish and Kamal Ndousse and Oliver Rausch and Nicholas Schiefer and Da Yan and Miranda Zhang and Ethan Perez},
journal={The Twelfth International Conference on Learning Representations},
year={2024},
note={\url{https://openreview.net/forum?id=tvhaxkMKAn}}
}

@article{perez-etal-2023-discovering,
    title = "Discovering Language Model Behaviors with Model-Written Evaluations",
    author = "Perez, Ethan  and
      Ringer, Sam  and
      Lukosiute, Kamile  and
      Nguyen, Karina  and
      Chen, Edwin  and
      Heiner, Scott  and
      Pettit, Craig  and
      Olsson, Catherine  and
      Kundu, Sandipan  and
      Kadavath, Saurav  and
      Jones, Andy  and
      Chen, Anna  and
      Mann, Benjamin  and
      Israel, Brian  and
      Seethor, Bryan  and
      McKinnon, Cameron  and
      Olah, Christopher  and
      Yan, Da  and
      Amodei, Daniela  and
      Amodei, Dario  and
      Drain, Dawn  and
      Li, Dustin  and
      Tran-Johnson, Eli  and
      Khundadze, Guro  and
      Kernion, Jackson  and
      Landis, James  and
      Kerr, Jamie  and
      Mueller, Jared  and
      Hyun, Jeeyoon  and
      Landau, Joshua  and
      Ndousse, Kamal  and
      Goldberg, Landon  and
      Lovitt, Liane  and
      Lucas, Martin  and
      Sellitto, Michael  and
      Zhang, Miranda  and
      Kingsland, Neerav  and
      Elhage, Nelson  and
      Joseph, Nicholas  and
      Mercado, Noemi  and
      DasSarma, Nova  and
      Rausch, Oliver  and
      Larson, Robin  and
      McCandlish, Sam  and
      Johnston, Scott  and
      Kravec, Shauna  and
      El Showk, Sheer  and
      Lanham, Tamera  and
      Telleen-Lawton, Timothy  and
      Brown, Tom  and
      Henighan, Tom  and
      Hume, Tristan  and
      Bai, Yuntao  and
      Hatfield-Dodds, Zac  and
      Clark, Jack  and
      Bowman, Samuel R.  and
      Askell, Amanda  and
      Grosse, Roger  and
      Hernandez, Danny  and
      Ganguli, Deep  and
      Hubinger, Evan  and
      Schiefer, Nicholas  and
      Kaplan, Jared",
    journal = "Findings of the Association for Computational Linguistics: ACL 2023",
    year = "2023",
    doi = "10.18653/v1/2023.findings-acl.847",
    pages = "13387--13434"
}

@article{sonderen_ineffectiveness_2013,
	title = {Ineffectiveness of Reverse Wording of Questionnaire Items: Let’s Learn from Cows in the Rain},
	volume = {8},
	issn = {1932-6203},
	shorttitle = {Ineffectiveness of {Reverse} {Wording} of {Questionnaire} {Items}},
	doi = {10.1371/journal.pone.0068967},
	language = {en},
	number = {7},
	journal = {PLoS ONE},
	author = {Sonderen, Eric Van and Sanderman, Robbert and Coyne, James C.},
	editor = {Baradaran, Hamid Reza},
	year = {2013},
	pages = {e68967},
}

@article{lejuez_evaluation_2002,
	title = {Evaluation of a behavioral measure of risk taking: The {Balloon Analogue Risk Task} ({BART}).},
	volume = {8},
	issn = {1939-2192, 1076-898X},
	shorttitle = {Evaluation of a behavioral measure of risk taking},
	doi = {10.1037/1076-898X.8.2.75},
	language = {en},
	number = {2},
	journal = {Journal of Experimental Psychology: Applied},
	author = {Lejuez, C. W. and Read, Jennifer P. and Kahler, Christopher W. and Richards, Jerry B. and Ramsey, Susan E. and Stuart, Gregory L. and Strong, David R. and Brown, Richard A.},
	year = {2002},
	pages = {75--84},
}

@article{DOSPERT_scale, 
title={A Domain-Specific Risk-Taking ({DOSPERT}) scale for adult populations}, 
volume={1}, 
DOI={10.1017/S1930297500000334}, 
number={1}, 
journal={Judgment and Decision Making}, 
author={Blais, Ann-Renée and Weber, Elke U.}, 
year={2006}, 
pages={33–47}
}

@article{cronbach_further_1950,
	title = {Further evidence on response sets and test design},
	volume = {10},
	copyright = {https://journals.sagepub.com/page/policies/text-and-data-mining-license},
	issn = {0013-1644, 1552-3888},
	doi = {10.1177/001316445001000101},
	language = {en},
	number = {1},
	journal = {Educational and Psychological Measurement},
	author = {Cronbach, Lee J.},
	year = {1950},
	pages = {3--31},
}

@article{couch_yeasayers_1960,
	title = {Yeasayers and naysayers: Agreeing response set as a personality variable.},
	volume = {60},
	issn = {0096-851X},
	shorttitle = {Yeasayers and naysayers},
	doi = {10.1037/h0040372},
	language = {en},
	number = {2},
	journal = {The Journal of Abnormal and Social Psychology},
	author = {Couch, Arthur and Keniston, Kenneth},
	year = {1960},
	pages = {151--174},
}

@article{jia_decision-making_2024,
	title = {Decision-making behavior evaluation framework for {LLMs} under uncertain context},
	volume = {37},
	doi = {10.52202/079017-3601},
	journal = {Advances in {Neural} {Information} {Processing} {Systems}},
	author = {Jia, Jingru and Yuan, Zehua and Pan, Junhao and McNamara, Paul and Chen, Deming},
	year = {2024},
	pages = {113360--113382},
}

@misc{openai2026scaling,
  author       = {{OpenAI}},
  title        = {Scaling {AI} for Everyone},
  year         = {2026},
  month        = feb,
  day          = {27},
  howpublished = {\url{https://openai.com/index/scaling-ai-for-everyone/}},
  note         = {Accessed May 2026}
}

@misc{askell2026constitution,
  author       = {Askell, Amanda and Carlsmith, Joe and Olah, Chris and Kaplan, Jared and Karnofsky, Holden and others},
  title        = {Claude's {C}onstitution},
  year         = {2026},
  month        = jan,
  institution  = {Anthropic},
  howpublished = {\url{https://www.anthropic.com/constitution}},
  note         = {Accessed May 2026}
}

@misc{openai2025modelspec,
  author       = {{OpenAI}},
  title        = {Model {S}pec},
  year         = {2025},
  month        = feb,
  howpublished = {\url{https://model-spec.openai.com/2025-02-12.html}},
  note         = {Accessed May 2026}
}

@article{baumgartner-steenkamp,
author = {Hans Baumgartner and Jan-Benedict E.M. Steenkamp},
title ={Response Styles in Marketing Research: A Cross-National Investigation},
journal = {Journal of Marketing Research},
volume = {38},
number = {2},
pages = {143-156},
year = {2001},
doi = {10.1509/jmkr.38.2.143.18840},
}

@article{ouyang2022training,
 author = {Ouyang, Long and Wu, Jeffrey and Jiang, Xu and Almeida, Diogo and Wainwright, Carroll and Mishkin, Pamela and Zhang, Chong and Agarwal, Sandhini and Slama, Katarina and Ray, Alex and Schulman, John and Hilton, Jacob and Kelton, Fraser and Miller, Luke and Simens, Maddie and Askell, Amanda and Welinder, Peter and Christiano, Paul F and Leike, Jan and Lowe, Ryan},
 journal = {Advances in Neural Information Processing Systems},
 pages = {27730--27744},
 title = {Training language models to follow instructions with human feedback},
 url = {https://proceedings.neurips.cc/paper\_files/paper/2022/file/b1efde53be364a73914f58805a001731-Paper-Conference.pdf},
 volume = {35},
 year = {2022}
}

@article{shu2024you,
    title = "You don{'}t need a personality test to know these models are unreliable: Assessing the Reliability of Large Language Models on Psychometric Instruments",
    author = "Shu, Bangzhao  and
      Zhang, Lechen  and
      Choi, Minje  and
      Dunagan, Lavinia  and
      Logeswaran, Lajanugen  and
      Lee, Moontae  and
      Card, Dallas  and
      Jurgens, David",
    journal = "Proceedings of the 2024 Conference of the North American Chapter of the Association for Computational Linguistics: Human Language Technologies (Volume 1: Long Papers)",
    year = "2024",
    doi = "10.18653/v1/2024.naacl-long.295",
    pages = "5263--5281",
}

@article{horton2023large,
  title={Large language models as simulated economic agents: What can we learn from homo silicus?},
  author={Horton, John J and Filippas, Apostolos and Manning, Benjamin S},
  year={2023},
  journal={National Bureau of Economic Research},
  doi = {10.3386/w31122},
  type = "Working Paper",
  series = "Working Paper Series",
}

@article{argyle_out_2023,
	title = {Out of one, many: Using language models to simulate human samples},
	volume = {31},
	copyright = {https://www.cambridge.org/core/terms},
	issn = {1047-1987, 1476-4989},
	shorttitle = {Out of {One}, {Many}},
	doi = {10.1017/pan.2023.2},
	language = {en},
	number = {3},
	journal = {Political Analysis},
	author = {Argyle, Lisa P. and Busby, Ethan C. and Fulda, Nancy and Gubler, Joshua R. and Rytting, Christopher and Wingate, David},
	year = {2023},
	pages = {337--351},
}

@article{lohn2024machine,
    title = "Is Machine Psychology here? On Requirements for Using Human Psychological Tests on Large Language Models",
    author = {L{\"o}hn, Lea  and
      Kiehne, Niklas  and
      Ljapunov, Alexander  and
      Balke, Wolf-Tilo},
    journal = "Proceedings of the 17th International Natural Language Generation Conference",
    year = "2024",
    doi = "10.18653/v1/2024.inlg-main.19",
    pages = "230--242"
}

@article{suhr2025stop,
      title={Position: Stop Evaluating {AI} with Human Tests, Develop Principled, {AI}-specific Tests instead}, 
      author={Tom S{\"u}hr and Florian E. Dorner and Olawale Salaudeen and Augustin Kelava and Samira Samadi},
      journal={arXiv preprint arXiv:2507.23009},
      year={2026},
      doi={10.48550/arXiv.2507.23009}, 
}

@article{gupta-etal-2024-self,
    title = "Self-Assessment Tests are Unreliable Measures of {LLM} Personality",
    author = "Gupta, Akshat  and
      Song, Xiaoyang  and
      Anumanchipalli, Gopala",
    journal = "Proceedings of the 7th BlackboxNLP Workshop: Analyzing and Interpreting Neural Networks for NLP",
    year = "2024",
    doi = "10.18653/v1/2024.blackboxnlp-1.20",
    pages = "301--314"
}

@article{pellert2024ai,
  title={{AI} psychometrics: Assessing the psychological profiles of large language models through psychometric inventories},
  author={Pellert, Max and Lechner, Clemens M and Wagner, Claudia and Rammstedt, Beatrice and Strohmaier, Markus},
  journal={Perspectives on Psychological Science},
  volume={19},
  number={5},
  pages={808--826},
  year={2024},
  publisher={Sage Publications Sage CA: Los Angeles, CA},
  doi={10.1177/17456916231214460}
}

@article{jiang2023evaluating,
 author = {Jiang, Guangyuan and Xu, Manjie and Zhu, Song-Chun and Han, Wenjuan and Zhang, Chi and Zhu, Yixin},
 journal = {Advances in Neural Information Processing Systems},
 pages = {10622--10643},
 title = {Evaluating and Inducing Personality in Pre-trained Language Models},
 url = {https://proceedings.neurips.cc/paper\_files/paper/2023/file/21f7b745f73ce0d1f9bcea7f40b1388e-Paper-Conference.pdf},
 volume = {36},
 year = {2023}
}

@article{hartley2025personality,
    title = "How Personality Traits Shape {LLM} Risk-Taking Behaviour",
    author = "Hartley, John  and
      Hamill, Conor Brian  and
      Seddon, Dale  and
      Batra, Devesh  and
      Okhrati, Ramin  and
      Khraishi, Raad",
    journal = "Findings of the Association for Computational Linguistics: ACL 2025",
    year = "2025",
    doi = "10.18653/v1/2025.findings-acl.1085",
    pages = "21068--21092"
}

@article{eberhardt_development_2025,
	title = {Development and validation of large language model rating scales for automatically transcribed psychological therapy sessions},
	volume = {15},
	issn = {2045-2322},
	doi = {10.1038/s41598-025-14923-y},
	number = {1},
	journal = {Scientific Reports},
	author = {Eberhardt, Steffen T. and Vehlen, Antonia and Schaffrath, Jana and Schwartz, Brian and Baur, Tobias and Schiller, Dominik and Hallmen, Tobias and André, Elisabeth and Lutz, Wolfgang},
	year = {2025},
	pages = {29541},
}

@article{faulborn-etal-2025-little,
    title = "Only a Little to the Left: A Theory-grounded Measure of Political Bias in Large Language Models",
    author = "Faulborn, Mats  and
      Sen, Indira  and
      Pellert, Max  and
      Spitz, Andreas  and
      Garcia, David",
    journal = "Proceedings of the 63rd Annual Meeting of the Association for Computational Linguistics",
    year = "2025",
    doi = "10.18653/v1/2025.acl-long.1529",
    pages = "31684--31704",
}

@article{kriegmair_machine_2026,
	title = {Machine individuality: Separating genuine idiosyncrasy from response bias in large language models},
	shorttitle = {Machine individuality},
	doi = {10.48550/arXiv.2604.16755},
	language = {en},
	author = {Kriegmair, Valentin and Wulff, Dirk U.},
	year = {2026},
	journal = {arXiv preprint arXiv:2604.16755},
}

@article{peereboom_cognitive_2025,
	title = {Cognitive phantoms in large language models through the lens of latent variables},
	volume = {4},
	issn = {29498821},
	doi = {10.1016/j.chbah.2025.100161},
	language = {en},
	journal = {Computers in Human Behavior: Artificial Humans},
	author = {Peereboom, Sanne and Schwabe, Inga and Kleinberg, Bennett},
	year = {2025},
	pages = {100161},
}

@article{jung-etal-2026-psychometric,
    title = "Do Psychometric Tests Work for Large Language Models? Evaluation of Tests on Sexism, Racism, and Morality",
    author = "Jung, Jana  and
      Lutz, Marlene  and
      Sen, Indira  and
      Strohmaier, Markus",
    journal = "Proceedings of the 19th Conference of the {E}uropean Chapter of the {A}ssociation for {C}omputational {L}inguistics (Volume 1: Long Papers)",
    year = "2026",
    doi = "10.18653/v1/2026.eacl-long.380",
    pages = "8143--8173",
}

@article{serapio-garcia_psychometric_2025,
	title = {A psychometric framework for evaluating and shaping personality traits in large language models},
	volume = {7},
	issn = {2522-5839},
	doi = {10.1038/s42256-025-01115-6},
	language = {en},
	number = {12},
	journal = {Nature Machine Intelligence},
	author = {Serapio-García, Gregory and Safdari, Mustafa and Crepy, Clément and Sun, Luning and Fitz, Stephen and Romero, Peter and Abdulhai, Marwa and Faust, Aleksandra and Matarić, Maja},
	year = {2025},
	pages = {1954--1968},
}

@article{JOHNSON201478,
title = {Measuring thirty facets of the {Five Factor Model} with a 120-item public domain inventory: Development of the {IPIP-NEO-120}},
journal = {Journal of Research in Personality},
volume = {51},
pages = {78-89},
year = {2014},
issn = {0092-6566},
doi = {10.1016/j.jrp.2014.05.003},
author = {John A. Johnson}
}

@article{frey_risk_2017,
	title = {Risk preference shares the psychometric structure of major psychological traits},
	volume = {3},
	issn = {2375-2548},
	doi = {10.1126/sciadv.1701381},
	language = {en},
	number = {10},
	journal = {Science Advances},
	author = {Frey, Renato and Pedroni, Andreas and Mata, Rui and Rieskamp, Jörg and Hertwig, Ralph},
	year = {2017},
	pages = {e1701381},
}

@article{willard2023efficient,
      title={Efficient Guided Generation for Large Language Models}, 
      author={Brandon T. Willard and Rémi Louf},
      year={2023},
      journal={arXiv preprint arXiv:2307.09702},
      doi={10.48550/arXiv.2307.09702}, 
}

@article{weijters_misresponse_2012,
	title = {Misresponse to reversed and negated items in surveys: A review},
	volume = {49},
	issn = {0022-2437, 1547-7193},
	shorttitle = {Misresponse to {Reversed} and {Negated} {Items} in {Surveys}},
	doi = {10.1509/jmr.11.0368},
	language = {en},
	number = {5},
	journal = {Journal of Marketing Research},
	author = {Weijters, Bert and Baumgartner, Hans},
	year = {2012},
	pages = {737--747},
}

@article{garcia-ferrero_this_2023,
  title={This is not a dataset: A large negation benchmark to challenge large language models},
  author={Garc{\'\i}a-Ferrero, Iker and Altuna, Bego{\~n}a and Alvez, Javier and Gonzalez-Dios, Itziar and Rigau, German},
  journal={Proceedings of the 2023 conference on empirical methods in natural language processing},
  pages={8596--8615},
  year={2023},
  doi={10.18653/v1/2023.emnlp-main.531}
}

@article{sun2026friendly,
author = {Sun, Yuan and Wang, Ting},
title = {Be Friendly, Not Friends: How {LLM} Sycophancy Shapes User Trust},
year = {2026},
doi = {10.1145/3772318.3791079},
journal = {Proceedings of the 2026 CHI Conference on Human Factors in Computing Systems},
articleno = {1576}
}

@article{kadambi2026anthropomorphism,
  title={Anthropomorphism and Trust in Human-Large Language Model interactions},
  author={Kadambi, Akila and D'Elia, Ylenia and Shah, Tanishka and Comsa, Iulia and Lentz, Alison and Siri-Ngammuang, Katie and Buechler, Tara and Kaplan, Jonas and Damasio, Antonio and Narayanan, Srini and others},
  journal={arXiv preprint arXiv:2604.15316},
  year={2026},
  doi={10.48550/arXiv.2604.15316},
}

@article{fitz2025psychometric,
  title={Psychometric Personality Shaping Modulates Capabilities and Safety in Language Models},
  author={Fitz, Stephen and Romero, Peter and Basart, Steven and Chen, Sipeng and Hernandez-Orallo, Jose},
  journal={arXiv preprint arXiv:2509.16332},
  year={2025},
  doi={10.48550/arXiv.2509.16332}, 
}

@article{cronbach1951coefficient,
  title={Coefficient alpha and the internal structure of tests},
  author={Cronbach, Lee J},
  journal={Psychometrika},
  DOI={10.1007/BF02310555},
  volume={16},
  number={3},
  pages={297--334},
  year={1951},
  publisher={Springer-Verlag}
}

@article{wulff2018meta,
  title={A meta-analytic review of two modes of learning and the description-experience gap.},
  author={Wulff, Dirk U and Mergenthaler-Canseco, Max and Hertwig, Ralph},
  journal={Psychological bulletin},
  volume={144},
  number={2},
  pages={140},
  year={2018},
  publisher={American Psychological Association},
  doi={10.1037/bul0000115}
}

@article{marsh1996positive,
  title={Positive and negative global self-esteem: A substantively meaningful distinction or artifactors?},
  author={Marsh, Herbert W},
  journal={Journal of personality and social psychology},
  volume={70},
  number={4},
  pages={810},
  year={1996},
  publisher={American Psychological Association},
  doi={10.1037/0022-3514.70.4.810}
}

@article{schmitt1985factors,
  title={Factors defined by negatively keyed items: The result of careless respondents?},
  author={Schmitt, Neal and Stults, Daniel M},
  journal={Applied Psychological Measurement},
  volume={9},
  number={4},
  pages={367--373},
  year={1985},
  doi = {10.1177/014662168500900405},
  publisher={Sage Publications Sage CA: Thousand Oaks, CA}
}

@article{krefeldtraining,
  title={Training data limits the prediction of consumer heterogeneity by {LLM}-based digital twins},
  author={Krefeld-Schwalb, Antonia and Johnson, Eric and Weaver, Caroline and Wulff, Dirk U},
  journal={OSF Preprints},
  year={2026},
  doi={10.31234/osf.io/97dya\_v1},
  publisher={OSF}
}

@article{kaiser2022scientific,
  title={The scientific value of numerical measures of human feelings},
  author={Kaiser, Caspar and Oswald, Andrew J},
  journal={Proceedings of the National Academy of Sciences},
  volume={119},
  number={42},
  pages={e2210412119},
  year={2022},
  publisher={National Academy of Sciences},
    doi = {10.1073/pnas.2210412119},
}

@article{kumar2026failure,
  title={Failure of contextual invariance in gender inference with large language models},
  author={Kumar, Sagar and Flint, Ariel and Aiello, Luca Maria and Baronchelli, Andrea},
  journal={arXiv preprint arXiv:2603.23485},
  year={2026},
  doi={10.48550/arXiv.2603.23485}
}

@article{binz2026post,
  title={Post-training makes large language models less human-like},
  author={Binz, Marcel and Akata, Elif and Almaatouq, Abdullah and Alsobay, Mohammed and Ariasov, Oleksii and Br{\"a}ndle, Franziska and Broska, David and Burton, Jason W and Busch, Nuno and Callaway, Frederick and others},
  journal={arXiv preprint arXiv:2605.07632},
  year={2026},
  doi={10.48550/arXiv.2605.07632}
}

@article{kuribayashi2024psychometric,
    title = "Psychometric Predictive Power of Large Language Models",
    author = "Kuribayashi, Tatsuki  and
      Oseki, Yohei  and
      Baldwin, Timothy",
    journal = "Findings of the Association for Computational Linguistics: NAACL 2024",
    year = "2024",
    doi = "10.18653/v1/2024.findings-naacl.129",
    pages = "1983--2005"
}

@article{hussain2025rebuttal,
  title={A rebuttal of two common deflationary stances against {LLM} cognition},
  author={Hussain, Zak and Mata, Rui and Wulff, Dirk U},
  journal={Findings of the Association for Computational Linguistics},
  pages={24208--24213},
  year={2025},
  doi={10.18653/v1/2025.findings-acl.1242}
}

@article{wulff2024behavioral,
  title={The behavioral and social sciences need open {LLMs}},
  author={Wulff, Dirk U and Hussain, Zak and Mata, Rui},
  journal={OSF Preprints},
  year={2024},
  doi={10.31219/osf.io/ybvzs}
}

@article{abdulhai2024moral,
    title = "Moral Foundations of Large Language Models",
    author = "Abdulhai, Marwa  and
      Serapio-Garc{\'i}a, Gregory  and
      Crepy, Clement  and
      Valter, Daria  and
      Canny, John  and
      Jaques, Natasha",
    journal = "Proceedings of the 2024 Conference on Empirical Methods in Natural Language Processing",
    year = "2024",
    doi = "10.18653/v1/2024.emnlp-main.982",
    pages = "17737--17752",
}

@article{rozado2024political,
  title={The political preferences of {LLMs}},
  author={Rozado, David},
  journal={PloS one},
  volume={19},
  number={7},
  pages={e0306621},
  year={2024},
  publisher={Public Library of Science},
  doi={10.1371/journal.pone.0306621}
}

@article{ye2025large,
  title={Large language model psychometrics: A systematic review of evaluation, validation, and enhancement},
  author={Ye, Haoran and Jin, Jing and Xie, Yuhang and Zhang, Xin and Song, Guojie},
  journal={arXiv preprint arXiv:2505.08245},
  year={2025},
  doi={10.48550/arXiv.2505.08245}
}

@article{xie2026aipsychobench,
  title={{AIPsychoBench}: Understanding the Psychometric Differences between {LLMs} and Humans},
  author={Xie, Wei and Wang, Zhenhua and Ma, Shuoyoucheng and Sun, Xiaobing and Chen, Kai and Wang, Enze and Liu, Wei and Tong, Hanying},
  journal={Topics in Cognitive Science},
  volume={18},
  number={2},
  pages={e70041},
  year={2026},
  doi = {10.1111/tops.70041},
  publisher={Wiley Online Library}
}

@article{tjuatja2024llms,
  title={Do {LLMs} exhibit human-like response biases? A case study in survey design},
  author={Tjuatja, Lindia and Chen, Valerie and Wu, Tongshuang and Talwalkar, Ameet and Neubig, Graham},
  journal={Transactions of the Association for Computational Linguistics},
  volume={12},
  pages={1011--1026},
  year={2024},
  issn = {2307-387X},
  doi = {10.1162/tacl\_a\_00685},
  publisher={MIT Press 255 Main Street, 9th Floor, Cambridge, Massachusetts 02142, USA~…}
}

@article{hussainhuman,
  title={How human-like is {LLM} cognition?},
  journal={OSF Preprints},
  author={Hussain, Zak and Mata, Rui and Wulff, Dirk U},
  year={2026},
  publisher={OSF},
  doi={10.31234/osf.io/2yvnt\_v2}
}

@article{xu2025large,
  title={Large language models without grounding recover non-sensorimotor but not sensorimotor features of human concepts},
  author={Xu, Qihui and Peng, Yingying and Nastase, Samuel A and Chodorow, Martin and Wu, Minghua and Li, Ping},
  journal={Nature human behaviour},
  volume={9},
  number={9},
  pages={1871--1886},
  issn = {2397-3374},
  doi = {10.1038/s41562-025-02203-8},
  year={2025},
  publisher={Nature Publishing Group UK London}
}

@article{li2006using,
  title={Using the BIDR to distinguish the effects of impression management and self-deception on the criterion validity of personality measures: A meta-analysis},
  author={Li, Andrew and Bagger, Jessica},
  journal={International Journal of Selection and Assessment},
  volume={14},
  number={2},
  pages={131--141},
  year={2006},
  doi = {10.1111/j.1468-2389.2006.00339.x},
  publisher={Wiley Online Library}
}

@article{goldberg1999broad,
  title={A broad-bandwidth, public domain, personality inventory measuring the lower-level facets of several five-factor models},
  author={Goldberg, Lewis R},
  journal={Personality psychology in Europe},
  volume={7},
  number={1},
  pages={7--28},
  year={1999},
  publisher={Tilburg Netherland}
}

@misc{johnson_ipipneo_data,
  author       = {Johnson, John A.},
  title        = {{IPIP-NEO} Data Repository},
  year         = {2014},
  howpublished = {Open Science Framework},
  note         = {\url{https://osf.io/tbmh5/}}
}

\end{document}